\pdfoutput=1 

\documentclass[11pt,a4paper]{article}
\usepackage[hyperref]{emnlp2020}
\usepackage{times}
\usepackage{latexsym}

\usepackage{booktabs} 
\usepackage{multirow}
\usepackage{microtype}
\usepackage{arydshln}
\usepackage{amsmath}
\usepackage{mathtools,amssymb}
\usepackage{csquotes}
\usepackage{cleveref}
\usepackage{caption}
\usepackage{subcaption}
\usepackage[english]{babel}
\usepackage{blindtext}
\usepackage{array}

\crefformat{section}{\S#2#1#3}
\crefformat{subsection}{\S#2#1#3}
\crefformat{subsubsection}{\S#2#1#3}
\crefrangeformat{section}{\S\S#3#1#4 to~#5#2#6}
\crefmultiformat{section}{\S\S#2#1#3}{ and~#2#1#3}{, #2#1#3}{ and~#2#1#3}
\aclfinalcopy 

\newcounter{example}[section]

\usepackage{enumitem}

\definecolor{mbc}{RGB}{255, 40, 40}
\definecolor{ncc}{RGB}{0, 153, 51}
\definecolor{esc}{RGB}{0, 40, 255}
\definecolor{flc}{RGB}{255, 140, 0}

\newcommand\blfootnote[1]{%
  \begingroup
  \renewcommand\thefootnote{}\footnote{#1}%
  \addtocounter{footnote}{-1}%
  \endgroup
}

\title{COMETA: A Corpus for Medical Entity Linking in the Social Media}

\author{Marco Basaldella$^{\dag,*}$, Fangyu Liu$^{\dag,*}$, Ehsan Shareghi$^{\dag,\ddagger}$, Nigel Collier$^\dag$ \\
  $^\dag$Language Technology Lab, University of Cambridge \\
  $^\ddagger$ University College London \\
  $^\dag$\texttt{\{mb2313, fl399, nhc30\}@cam.ac.uk}\\ $^\ddagger$\texttt{e.shareghi@ucl.ac.uk}
}

\date{}

\begin{document}
\maketitle
\begin{abstract}


Whilst there has been growing progress in Entity Linking (EL) for general language, existing datasets fail to address the complex nature of health terminology in layman's language. Meanwhile, there is a growing need for applications that can understand the public's voice in the health domain. To address this we introduce a new corpus called COMETA, consisting of 20k English biomedical entity mentions from Reddit expert-annotated with links to SNOMED CT, a widely-used medical knowledge graph. Our corpus satisfies a combination of desirable properties, from scale and coverage to diversity and quality, that to the best of our knowledge has not been met by any of the existing resources in the field. Through benchmark experiments on 20 EL baselines from string- to neural-based models we shed light on the ability of these systems to perform complex inference on entities and concepts under 2 challenging evaluation scenarios. Our experimental results on COMETA illustrate that no golden bullet exists and even the best mainstream techniques still have a significant performance gap to fill, while the best solution relies on combining different views of data.

\end{abstract}

\section{Introduction}
\label{sec:intro}
Social media has become a dominant means for users to share their opinions, emotions and daily experience of life. A large body of work has shown that informal exchanges such as online forums can be leveraged to supplement traditional approaches to a broad range of public health questions such as monitoring suicidal risk and depression \cite{benton2017multitask}, domestic abuse~\cite{schrading-etal-2015-analysis}, cancer \cite{nzali2017patients}, and epidemics \cite{aramaki2011twitter,DBLP:journals/csur/JoshiKSPM20}.\blfootnote{$^*$Equal contribution.} 





One of the widely exercised steps to establish a semantic understanding of social media is Entity Linking (EL), i.e., the task of linking { entities} within a text to a suitable concept in a reference Knowledge Graph (KG) \cite{liu2013entity,yang2015s,DBLP:conf/emnlp/YangCE16, DBLP:conf/www/RanSW18}.
However,
it is well-documented that poorly composed contexts, the ubiquitous presence of colloquialisms, shortened forms, typing/spelling mistakes, and out-of-vocabulary words introduce challenges for effective utilisation of social media text~\cite{baldwin2013noisy,michel2018mtnt}.

These challenges are exacerbated in EL for user generated content (UGC) 
in the health domain for two main reasons: lack of dedicated annotated resources for  training EL models, 
and entanglement of the aforementioned challenges in general social media with the inherent complexity of the health domain and its terminology~(see Table~\ref{table:intro1}). 

\begin{figure}
  \centering
  \begin{subfigure}[b]{0.48\textwidth}
    \includegraphics[width=\textwidth]{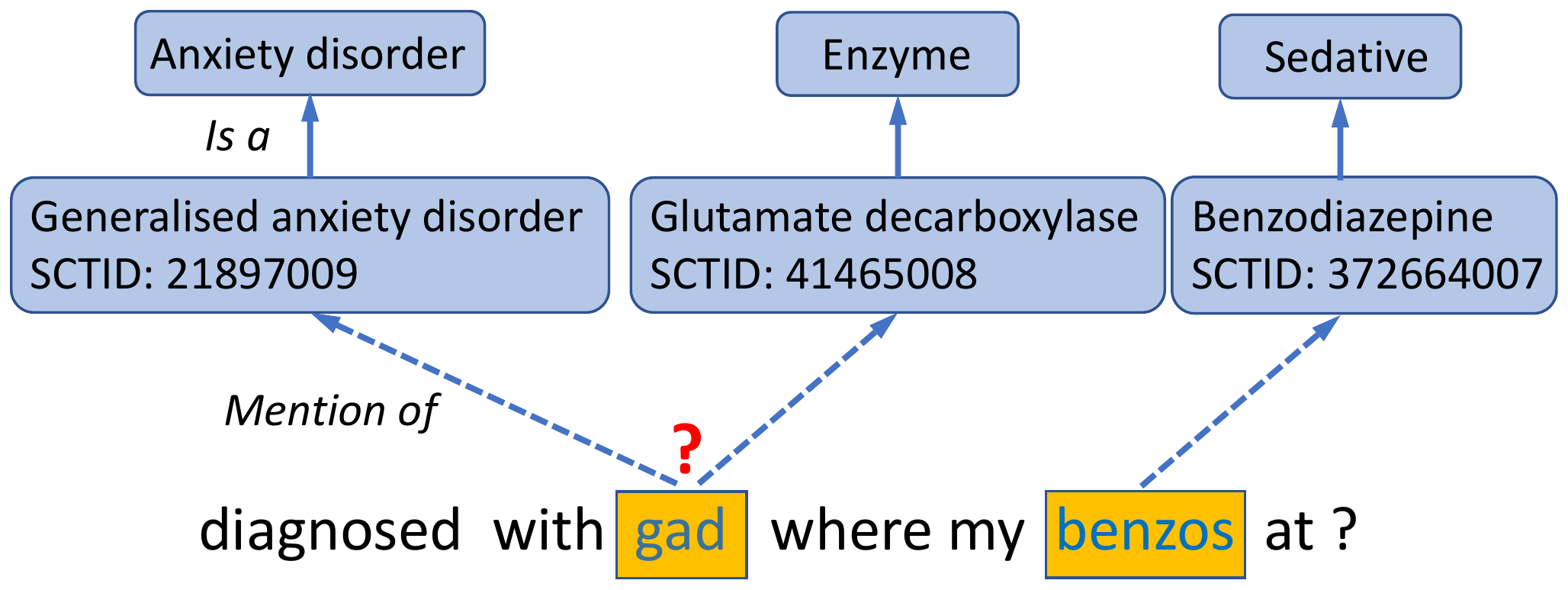}
    \caption{}
  \end{subfigure}
    \begin{subfigure}[b]{0.48\textwidth}
    \includegraphics[width=\textwidth]{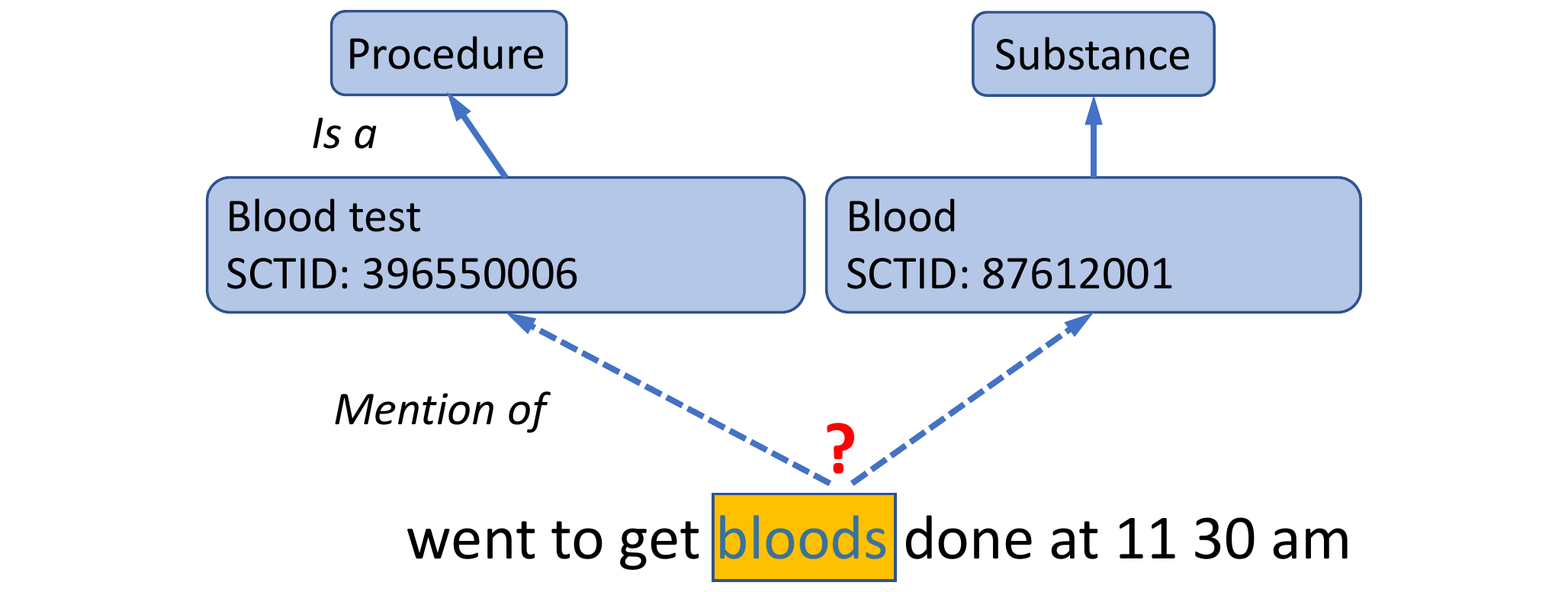}
    \caption{}
  \end{subfigure}
  \caption{Examples of the EL inference challenges for user generated text in the health domain.}
  \label{fig:example}
\end{figure}
%
\noindent 

\begin{table*}[ht]
\centering
\begin{tabular}{l l l}
\toprule
\textbf{Input term}  & \textbf{Gold SNOMED label} & \textbf{Challenge} \\ 
\midrule
{\it scratchy throat} & Pharyngeal dryness & {\it Colloquial symptom} \\
{\it lower right abdomen} & Structure of right lower quadrant of abdomen & {\it Term compositionality} \\
{\it anti nausea meds} & Medicinal product acting as antiemetic agent & {\it Negated term} \\
{\it MSM} & Dimethul sulfone & {\it Alternative product name} \\
{\it up all night cleaning} & Obsessive compulsive disorder & {\it Complex inference} \\
\bottomrule
\end{tabular}
\caption{Challenging examples of laymen's terms in COMETA and their target SNOMED concepts.}
\label{table:intro1}
\end{table*}
For example, in Figure~\ref{fig:example} we show sentences taken from social media where the semantics of the concept linking is complex and context-dependent.
In the first case, ``\emph{diagnosed with \underline{gad} where by \underline{benzos} at}'', \emph{benzos} is a colloquial form of benzodiazepines, a type of \emph{sedative}, and if  correctly resolved can provide a contextual clue to assign the appropriate sense to the polysemous term \emph{gad}: an abbreviation for \emph{generalised anxiety disorder} rather than e.g. \emph{glutamate decarboxylase}. In the second example, ``\emph{went to get \underline{bloods} done at 11 30}'', the word \emph{bloods} could be interpreted literally as \emph{blood}; however, in this case it clearly refers to a \emph{blood test}, and it can be correctly resolved only by considering the full context in which it is used.






In this paper we open up a new avenue for EL research specifically targeted at the important domain of health in social media through the release of a new resource: the  Corpus of Online Medical EnTities ({COMETA}), consisting of 20K biomedical entity mentions in English from publicly available and anonymous 
health discussions on Reddit. Each mention has been been expert-annotated with KG concepts\footnote{Throughout the paper {\it concept} refers to nodes in a KG (i.e., SNOMED), {\it term/entity} refers to the surface form mention of a concept in text, and {\it context} refers to the text in which a term appears. Also, SCTID denotes SNOMED CT Identifier.} from SNOMED CT \cite{donnelly2006snomed}\footnote{We use the July 2019 release of the international edition.}, a structured medical vocabulary of ca.350K concepts 
widely used to code Electronic Health Records (EHRs). As we show, COMETA provides a high quality yet challenging benchmark for developing EL techniques, especially for concepts not encountered during training ({\it zero-shot concepts}). Due to its semantic diversity the corpus represents an important pathway to knowledge integration between layman's language, EHRs and research evidence. 

Through a set of experiments we shed light on the challenges in this domain for several EL baselines utilising a diverse range of techniques from basic string-matching to low-dimensional entity embeddings \cite{bojanowski2017enriching}, KG structure embeddings \cite{grover2016node2vec,agarwal2019snomed2vec}, and context aware BERT embeddings \cite{devlin2019bert,btz682}. 
We show a simple augmentation of the mainstream BERT model with a Multi-Level Attention module can improve its effectiveness in capturing the contextual nuances of highly diverse layman's language in the health domain. Our experimental results illustrate that the best solution needs to combine multiple views of data and still heavily relies on basic techniques, while the remaining performance gap highlights the challenging nature of COMETA. We summarise these challenges and underline some of the key areas that are indispensable for further progress in this domain.

\section{Related Work and Datasets}
\label{sec:related}


\paragraph{Entity Linking.} EL \cite{bunescu2006using} is an important task that has sparked attention in recent years due to its wide-scale potential to aid in knowledge acquisition, e.g. the complementary problems of cross-document coreference resolution \cite{dredze2016twitter}, semantic relatedness \cite{dor2018semantic}, geo-coding \cite{gritta2017vancouver} and relation extraction \cite{koch2014type}. 

Systems that link entities to Wikipedia ({\it Wikification}) \cite{liu2013entity,roth2014wikification} and scientific literature to biomedical ontologies \cite{zheng2015entity} have been the focus of attention for many years. Generic EL systems such as Babelfy \cite{moro2014entity} and Tagme \cite{ferragina2011fast} identify and map entities to Wikipedia and WordNet \cite{miller1990introduction} but do not directly integrate the coding standards of healthcare KGs such as SNOMED. Medical EL systems such as cTAKES \cite{savova2010mayo} and MetaMap \cite{aronson2010overview} were designed to perform medical EL on EHRs but limited evidence e.g. \cite{denecke2014extracting} points to a large drop in recall on UGC such as patient forums. 


\paragraph{Medical EL in Social Media.} There are several medical EL corpora based on scientific publications \cite{pmid22901054,mohan2019medmentions}, EHRs \cite{suominen2013overview} and death certificates \cite{goeuriot2017clef}. However, none of these EL corpora dealt with the challenges of UGC. 

Due to under-reporting of drug side effects \cite{freifeld2014digital} pharmacovigilance datasets have been among the popular UGC benchmarks for evaluating medical EL. The earliest corpus in this domain was CADEC \cite{karimi2015cadec} where 1253 AskAPatient posts (6754 concept mentions) were annotated based on a search for the drugs Diclofenac and Lipitor. Another dataset, Twitter ADR \cite{nikfarjam2015pharmacovigilance}, consists of 1784 posts (1280 concept mentions) based on a search for 81 drug names, while TwiMed \cite{alvaro2017twimed} provides a comparable corpus of 1K PubMed and 1K Twitter texts (3144 concept mentions) based on a search for 30 drugs. \citet{limsopatham2016normalising} introduced two Twitter datasets (201 and 1436 concept mentions) with mappings to the SIDER-4 database~\cite{kuhn2016sider}, and 
RedMed \cite{LAVERTU2019103307} used Reddit to build a lexicon of alternative spellings for 2978 drugs to improve EL on social media.
Closest to our work is MedRed~\cite{scepanovic2020extracting}, a medical Named Entity Recognition corpus of 2K Reddit posts based on forums for 18 diseases. However we note several key differences to our work: our corpus is four times larger, provides two levels of mapping to general and context-specific concepts and has a much greater diversity of concepts rather than just symptoms and drugs~(\cref{sec:data-description}).

\section{The COMETA Corpus}
\label{sec:data}




The COMETA corpus satisfies multiple properties which we will explain throughout this section:
\begin{description}[noitemsep,leftmargin=2.5mm,topsep=0pt,parsep=0pt,partopsep=0pt]
\item[\textsc{Consistency.}]  COMETA has been annotated by biomedical experts to a high quality using SNOMED CT concepts (SCTIDs) - a standard for clinical information interchange~(\cref{sec:data-guideline});
\item[\textsc{Scale and Scope.}] To the best of our knowledge, with at 20K concept mentions, it is the largest UGC corpus for medical EL. Annotated entities cover a wide range of concepts including symptoms, diseases, anatomical expressions, chemicals, genes, devices and procedures across a range of conditions~(\cref{sec:data-description});
\item[\textsc{Distribution.}] We release the full corpus along with two sampling strategies ({\it Stratified} and {\it Zero-shot}) to prevent over-optimistic reporting of performance \cite{tutubalina2018medical}: while {\it Stratified} is designed to show the ability of systems to recognise known concepts with possibly novel mentions, {\it Zero-shot} is designed to test for recognising novel concepts~(\cref{sec:data-splits}). 
\end{description}



\subsection{Collection}
\label{sec:data-collection}
In order to build our corpus, we crawled health-themed forums on Reddit using Pushshift \cite{baumgartner2020pushshift} and Reddit's own APIs.
We choose forums satisfying strict constraints, i.e. selecting \emph{subreddits} where: (i) new content was posted daily, (ii) the quality of the content was sufficient (e.g. avoiding spam-ridden forums), (iii) the focus was the personal experiences or questions of the users.\footnote{For example, acceptable subreddits were \texttt{r/health}, \texttt{r/cancer}, \texttt{r/mentalhealth}, but not \texttt{r/medical\_news/}.} Applying these criteria, we selected a list of 68 subreddits (see Appendix~\ref{fullistreddits} for the full list) and crawled all the threads from 2015 to 2018, obtaining a collection of more than 800K discussions. This collection was then pruned by removing deleted posts, comments by bots or moderators, and so on.

In order to obtain the candidate entities, we trained the Flair NER system \cite{akbik2018coling} on a corpus of patient discussions from the health forum HealthUnlocked\footnote{The data for this system was provided by HealthUnlocked  (\url{https://healthunlocked.com/)} and cannot be publicly released in compliance with our data access agreement. The usage of this data was approved by the 
University of Cambridge's School of Humanities and Social Sciences Ethics Committee.
}; we then used this system to find medical entities in a random sub-sample of 100K discussions of our Reddit set, resulting in over 65K distinct named entities being discovered. 

Following the standard practices for ethical health research in social media outlined in \cite{benton-etal-2017-ethical}, we then anonymised the corpus to preserve, as far as possible, the privacy of the users. We removed personally identifiable data from messages and we selected terms that were mentioned by at least five users to avoid using terminology particular to a specific user.

Finally, after anonymisation, we hired two professional annotators with Ph.D. qualification in the biomedical domain to annotate the most popular 8K tagged entities 
with SNOMED concepts.  
%
%
\subsection{Consistency}
\label{sec:data-guideline}
The annotation process consisted of two steps: 
\begin{description}[noitemsep,leftmargin=2.5mm,topsep=0pt,parsep=0pt,partopsep=0pt]
\item[\textsc{First Step.}]We showed the first annotator an entity and up to six random sentences
in which it appeared.
If the entity was unambiguous, e.g. \emph{left ankle}, the annotator had to associate it to the relevant SCTID (e.g. SCTID: 51636004 -- \emph{Left Ankle}) and up to three sentences correctly representing it.
%
Moreover, the first annotator was required to mark NER system mistakes (e.g., wrong type, wrong span, or non-medical entity) to ensure the inclusion of high quality entities. 
Only 2.1\% of the entities were rejected, confirming the quality of our NER system. 

\item[\textsc{Second Step.}] The second annotator then tackled the ambiguous entities, 
selecting up to three possible \emph{specific} senses, and associating each sense to the relevant examples. 
This way, we obtained two levels of annotation: The \emph{General} level,  concerned with the literal meaning of the term, and the \emph{Specific} level, which takes into account the \emph{context} in which the entity appears. 
\end{description}

\noindent For example in the sentence ``\emph{Regarding my eyes, I'm not experiencing cloudiness.}'', the literal interpretation of the entity \emph{cloudiness} corresponds to the \emph{General} SNOMED concept SCTID: 81858005 -- \emph{Cloudy (qualifier value)}; however, a context-sensitive assignment which takes into account the word \emph{eyes} maps the entity to the  \emph{Specific} concept SCTID: 246636008 -- \emph{Hazy vision}. 
The \emph{specific} level requires contextual information to be effectively incorporated in the linking step, hence constitutes a  more challenging EL task.

The final corpus contains 20015 entities, each are assigned a \emph{General} and  \emph{Specific} SCTIDs and accompanied by an example sentence from Reddit where the entity is used. We also provide the link to the Reddit thread where the sentence appears (see Appendix~\ref{examplefromcometa} for a sample). Also, contrary to other corpora, we exclude NIL entities, i.e. entities without a corresponding concept in SNOMED.

\subsubsection{Assessing Annotation Quality}
\label{sec:annotator-performance}
Similar to \citet{mohan2019medmentions}, we assessed the quality of the annotation process by asking two pairs of  assessors\footnote{3 senior Ph.D. graduates and a PhD candidate in NLP. Note that there was no overlap between Annotators and Assessors.} to assess the quality of 1K random annotations (500 per pair of assessors). 

\paragraph{Assessor Guidelines.} We asked the assessors to evaluate the correctness of the expert assigned concepts on a discrete scale $[1,5]$, 1 being completely incorrect, and 5 being completely correct assignments. For example, mapping ``\emph{chronic back pain}'' in the sentence ``\emph{I have chronic low back pain.}'' to  SCTID: 134407002 -- \emph{Chronic back pain} entails a score of 5, to SCTID: 61968008 -- \emph{Syringe} entails a score of 1, and to SCTID: 77568009 -- \emph{Back} entails a score of 3, since the selected node is not correct but it identifies the location of the concept; see Table~\ref{table:assessor} in the Appendix~\ref{examplefromassessorguideline} for more details on the instructions we provided to the assessors.

\paragraph{Outcome.} Out of 1K examples, both assessors assigned the maximum score of 5 to 93.5\% and at least 4 to 96.8\% of both the \emph{general} and \emph{specific} level annotations. This is a good indication of the quality of the annotations and is in line with \citet{mohan2019medmentions}'s findings. Further investigation of weakly scored entities (3.2\% of examples) highlights the unique challenges that emerge in this domain. We provide two representative examples:

\begin{description}[noitemsep,leftmargin=2.5mm,topsep=0pt,parsep=0pt,partopsep=0pt]
\item[\textsc{Example 1.}] Regarding the entity ``\emph{UI}'' in the sentence ``\emph{If you're having GI problems, UI issues and/or ED issues please get the breath test for H.Pylori.}'', the annotator assigned the SCTID: 68566005 --  \emph{Urinary tract infectious disease}. One assessor agreed with the annotator's judgement on considering ``\emph{UI}'' as an abbreviation of ``\emph{Urinary infection}'', while the other assessor assigned only a score of 3, considering it as the abbreviation of ``\emph{Urinary incontinence}''. Given the sentence, however, both interpretations are plausible.

\item[\textsc{Example 2.}] Consider the entity ``\emph{pissed off}'' in the sentence ``\emph{And to top it off my stomach becomes bloated and pissed off.}''. Here, ``\emph{pissed off}'' is used figuratively to indicate some form of discomfort; however, the annotator assigned SCTID: 75408008 -- \emph{Feeling angry} which both assessors flagged as incorrect. Nevertheless, both assessors couldn't suggest a better SNOMED concept, as this phrase does not identify a precise disease.
\end{description}
These ambiguities exemplify why performing EL in the UGC domain can be hard even for humans and highlight the complexity found in laymen's medical conversations. 

\begin{figure}
  \centering
    \includegraphics[width=0.48\textwidth]{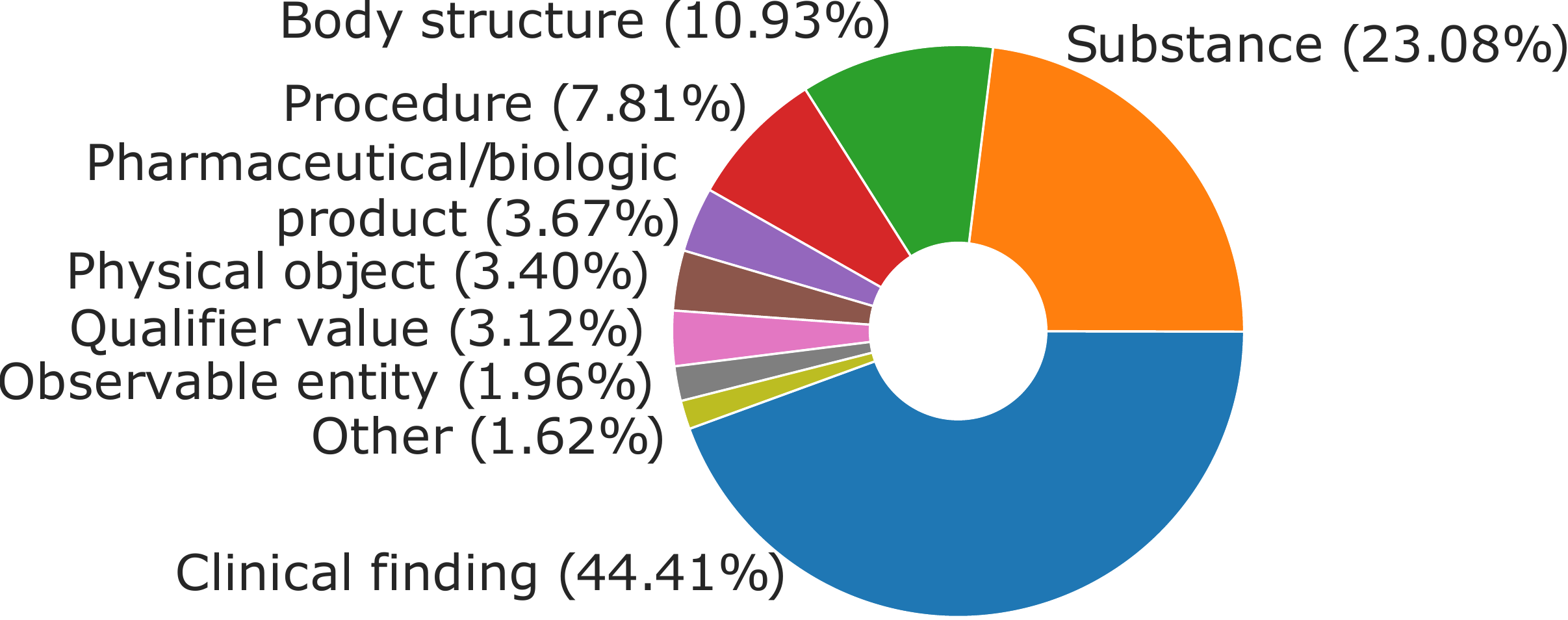}
      \caption{The semantic diversity of SNOMED concepts in COMETA.}
      \label{fig:distribution}
\end{figure}
\subsection{Scale and Scope}
\label{sec:data-description}
The corpus contains 6404 unique terms, 19911 unique example strings, 3645 unique \emph{general} concepts (SCTIDs), and 4003 unique  \emph{specific} concepts (SCTIDs). Each  \emph{general} and  \emph{specific}  concept is represented on average with more than 1 surface form, while some concepts had more than 15 surface forms, like for example SCTID: 5935008 -- \emph{Oral contraception},  SCTID: 225013001 -- \emph{Feeling bad}, and SCTID: 34000006 -- \emph{Chrohn's Disease}.

Additionally, each concept was accompanied by an average of at least 5 example sentences (median of 3), while 4.5\% of entities were linked to different \emph{general} and \emph{specific} SNOMED concepts (i.e., due to polysemy or contextual cues). We note that 31 entities are associated to more than one  \emph{general} SCTID, while 453 are associated to more than one  \emph{specific} SCTID. 

As illustrated in Figure~\ref{fig:distribution}, the most popular SNOMED domains in COMETA are
Clinical finding (44.4\%), Substance (23.1\%),
Body structure (10.9\%), Procedure (7.8\%), and Pharmaceutical / biologic product (3.7\%), covering more than 90\% of all the entities in the corpus (see Appendix~\ref{distributionofconceptsinstratifiedandzeroshot} for more details).

\subsection{Distribution}
\label{sec:data-splits}
We provide the COMETA corpus in two different sampled splits:
\begin{description}[noitemsep,leftmargin=2.5mm,topsep=0pt,parsep=0pt,partopsep=0pt]
\item[\textsc{Stratified Split.}]{Each SNOMED concept appearing in the test/development sets, appears at least once in the training set. The stratification by SCTID results in 100\% coverage of concepts in test/development, but on the surface form it covers only 58\% of the entities in the test set.}
\item[\textsc{Zero-Shot Split.}]{Development and test sets contain only novel concepts for which no training data was available.}
\end{description}

\noindent 
In other words, the Stratified split is designed to ensure that the model encounters the same concepts in the training, development and test set, but possibly with different surface forms;
the Zero-Shot split, instead, exposes models to unseen terms \emph{and} concepts in the development and testing sets, making it the hardest of the two settings (\cref{sec:experiments}). We argue that Zero-Shot is a more realistic setting since obtaining training data that covers all 350K SNOMED concepts involves a very expensive annotation effort. The statistics for the splits are shown in Table~\ref{table:dataset-counts}.


\section{Experiments and Results}
\label{sec:experiments}
In this section we conduct a diverse set of EL experiments, where we apply different simple and complex paradigms to link the annotated entities (and the sentences in which they appear) with the corresponding SNOMED concepts. 
We follow previous works in biomedical entity linking and use top-$k$ Accuracy ($k\in\{1,10\}$) to evaluate performance of EL systems \citep{dsouza-ng-2015-sieve}. Note that Acc@10 is only computed for systems returning a ranked list
%
and measures if the correct concept is contained within the top 10 concepts returned by the system. We also report Mean Reciprocal Rank (MRR, \citet{Craswell2018}), which instead measures the \emph{position} of the correct concept in the list of concepts returned by the system. Details about training as well as model and hardware configurations are 
available in Appendix~\ref{reproducibilitychecklist}.

\begin{table}[t]
\small
\centering
\begin{tabular}{@{}ll ccc@{}}
\toprule
                            &          & \textbf{Training} & \textbf{Dev} & \textbf{Test} \\ \midrule
\multirow{2}{*}{\textbf{Stratified}} & \textbf{General}  & 13489             & 2176         & 4350          \\
                            & \textbf{Specific} & 13441             & 2205         & 4369          \\
\multirow{2}{*}{\textbf{Zero-Shot}}  & \textbf{General}  & 14062             & 1958         & 3995          \\
                            & \textbf{Specific} & 13714             & 2018         & 4283          \\ \bottomrule
\end{tabular}
\caption{Number of examples in COMETA's splits.}
\label{table:dataset-counts}
\end{table}
 
Our baselines cover both string/dictionary-based algorithms (\cref{sec:string-dictionary}) which are good at capturing surface-level similarities, and neural models capable of incorporating contextual information (\cref{sec:neural-baselines}), where we experiment with a new Multi-Level Attention mechanism based on BERT to allow more efficient incorporation of context. Finally, to achieve the best possible performance, we combine these models in a back-off setting  where we leverage the benefits of each paradigm~(\cref{sec:back-off}). When describing the results, we will report the results on the general split and place the results on the specific split in parentheses. 

\subsection{Dictionary and String-based  Baselines}
\label{sec:string-dictionary}
As a first step, we experimented with a set of naïve systems based on string matching and edit distance.\footnote{For this set of experiments, we transform all entities and labels to lower-case.} 
These baselines ignore the context around the entities, since they
simply try to match entities against SNOMED labels. 

\paragraph{Dictionary.} A lookup table is built by traversing the training data, recording every entity and its corresponding SCTID, and
directly applied on the test set. 
If an entity is mapped to multiple SNOMED labels, the dictionary records the most frequent one. 

\paragraph{String-Matching Edit-Distance.} For every term, a string-matching search is conducted on its surface form against all the SNOMED node labels. Note that every SNOMED node has multiple alternative surface forms resulting in 2-36
comparisons per each entity. We count as a hit if the entity is matched with any of the node's surface forms based on exact match, Levenshtein ratio or Stolois distance, two strong string matching heuristics, which are defined as follows: given two strings $x,y$ the Levenshtein ratio (or normalised Levenshtein distance, \citet{yujian2007normalized}) is defined as 
$\frac{\text{Lev}(x, y)}{\max(|x|,|y|)}$
where Lev is the Levenshtein distance \cite{levenshtein1966binary} between $x$ and $y$; the Stoilos distance \citep{stoilos2005string} is defined as the similarity of two strings as $\texttt{comm}(x,y) - \texttt{diff}(x,y) + \texttt{winkler}(x,y)$
where the first and second terms are commonality and difference scores computed based on lengths of substrings of $x,y$ that are matched/unmatched 
and the third term is Jaro-Winkler distance \citep{winkler1999state}. Both edit distance metrics were tuned to offer the best trade-off between true and false positives in the development set; further details are provided in  Appendix~\ref{stoilosdistancedefinition}.

\begin{table}
\small
\setlength{\tabcolsep}{4pt}
\centering
\begin{tabular}{lccc}
\toprule
&&\multicolumn{2}{c}{{Acc@1}}\\
\cmidrule[1.5pt]{3-4}
\# & Method &Stratified Split & Zero-Shot Split\\
\midrule
s.1& Dictionary & .51 (.45) & 0 (0) \\
\midrule
s.2& Exact matching & .40 (.38) & .37 (.35)  \\
s.3 & Levenshtein ratio & .49 (.47) & .52 (.49) \\
s.4 & Stoilos distance & .51 (.49) & .53 (.51)\\
\midrule
s.5 & cTAKES & .51 (.48) & .53 (.47)\\
s.6 & QuickUMLS & .31 (.30) & .43 (.38) \\
\bottomrule
\end{tabular}
\caption{Comparison for Dictionary,  String-Matching, cTAKES and QuickUMLS baselines on stratified and zero-shot splits for general and (specific) levels.}
\label{Table:string_results}
\end{table}

\paragraph{cTAKES.} cTAKES \cite{savova2010mayo} is a heavily engineered system for processing clinical text. We report on its EL pipeline which is
based on several dictionary-based and advanced string matching techniques for resolving abbreviations, acronyms, spelling variants, and synonymy.\footnote{We also experimented with feeding the full text (including the entity) to cTAKES, but results were substantially worse.}

\paragraph{QuickUMLS.} QuickUMLS \cite{soldaini2016quickumls} is a fast approximate dictionary matching system for medical concept extraction using SimString \cite{Okazaki:Coling2010} as its back-end. We restrict its search space to the SNOMED CT subset of UMLS. As QuickUMLS predicts UMLS CUI instead of SCTID, we map predicted CUIs to SCTIDs through the UMLS api.\footnote{\url{https://documentation.uts.nlm.nih.gov/rest/home.html}} When multiple plausible mappings exist, we count a hit if anyone of them matches.\footnote{Unlike cTAKES, we found that feeding the full text to QuickUMLS yields slightly better results than using the entity only.}

\paragraph{Results.} Table~\ref{Table:string_results} summarises the results for the dictionary and string-based baselines. The dictionary method can serve as a strong baseline on the Stratified split, where its performance is barely matched by the more complex string-matching techniques. The most complex strategy, Stoilos distance, outperforms the other string-based techniques, and interestingly is on par with the highly complex cTAKES system while performing significantly better than QuickUMLS.
It is worth noting that cTAKES obtained 95.7\% in an EL task on an EHR dataset \cite{savova2010mayo}, highlighting the greater difficulty of the task when performed on the layman's language typical of UGC. 

Additionally, contrary to cTAKES, none of the string-based baselines are relying on external resources 
which might offer an improvement in resolving some abbreviations or acronyms that our string-based systems miss and cTAKES disambiguates correctly (e.g. ``\emph{ADHD}'' to SCTID: 406506008 -- \emph{Attention deficit hyperactivity disorder}). We leave further exploitation of such resources for future work.

\subsection{Neural-based Baselines}
\label{sec:neural-baselines}
For our neural setting, we define the problem as a cross-space mapping task by representing COMETA entities (along with their contexts) and SNOMED concepts using different text- and graph-based representation learning techniques, and then mapping the learned representations from the textual space to SNOMED concepts space.

\paragraph{Entity Embeddings.} 
We experimented both with ``traditional'' and contextual embedding techniques. To generate the entity embeddings we use FastText (FT, \citet{bojanowski2017enriching}) and BioBERT~\cite{btz682}, a PubMed-specialised version of BERT \cite{devlin2019bert}. The former was trained and the latter was further specialised on the set of 800K Reddit discussions described earlier~(\cref{sec:data-collection}).\footnote{Note that BERT here is used as a feature extractor. We tried finetuning BERT jointly with the alignment model, but performance got worse due to overfitting. We leave properly finetuned BERT models on COMETA as future work.}
In the case of multi-word terms, their embeddings were generated via averaging.\footnote{We tried replacing the entity embeddings with sentence embeddings via RNN/transformers, however, the performance was much worse. We speculate this was due to polluting the informative signal of an entity with its surrounding words. We leave further exploration of this to future work.} The dimensionality of the embeddings was 300 for FastText and 768 for BERT, and we denote them as FT-term and BERT-term, respectively.
Note that we acknowledge there are alternative options of BioBERT like SciBERT \citep{beltagy2019scibert} and ClinicalBERT \citep{alsentzer-etal-2019-publicly}. In our own experiments, we discovered that the further specialisation on Reddit discussions is more important than the choice of base model. That said, we leave explorations of other $\ast$BERT models on COMETA for future work.

\begin{table*}[t]
\small
\setlength{\tabcolsep}{6.8pt}
\centering
\begin{tabular}{lcccllllll}
\toprule
&&&\multicolumn{3}{c}{{Stratified Split}}&\multicolumn{3}{c}{{Zero-Shot Split}}\\
\cmidrule[1.5pt](lr){4-6}\cmidrule[1.5pt](lr){7-9}
\# &term embeddings & concept embeddings &   Acc@1 & Acc@10 & MRR&   Acc@1 & Acc@10 & MRR\\
\midrule
n.1&FT-term & FT-label & .40 (.38) & .71 (.70) & .51 (.49)& .21 (.20) & .53 (.51) & .31 (.30)\\
n.2&FT-term & node2vec &  .17 (.12) & .36 (.31) & .24 (.19)& .01 (.03) & .09 (.11) & .04 (.06)\\
n.3&BERT-term & BERT-label& .32 (.29) & .58 (.56)& .41 (.39) & .24 (.23) & .50 (.50) & .32 (.32)\\
n.4&$\text{BERT-term}_{\text{MLA}}$ & BERT-label& .38 (.35) & .66 (.63) & .48 (.45) & .29 (.27) & .56 (.52) & .38 (.35) \\
\midrule
n.5$^\mathbf{**}$&n.1 &   n.1 $\oplus$ n.2   & .47 (.42) & .76 (.73) & .57 (.49)& .12 (.12) & .37 (.41)  & .20 (.22) \\
n.6$^\mathbf{**}$&n.1$^\mathbf{*}$ $\oplus$ n.4& n.1 $\oplus$ n.2 $\oplus$ n.3 & .67 (.61) & .88 (.86) & .74 (.70) & .36 (.33) & .66 (.63) & .46 (.43)\\
\specialrule{.4em}{.3em}{.3em}
\multicolumn{9}{l}{\textbf{*} : A transformation is applied to FT-term ($[\ \mathbf{W}\cdot \text{FT-term}+\mathbf{b}\ ]_+$) before concatenation.}\\
\multicolumn{9}{l}{\textbf{**} : Alignment model used for these marked cases is just a linear transformation (without ReLU).}\\
\bottomrule
\end{tabular}
\caption{Comparison for neural-based baselines on stratified and zero-shot splits for general and (specific) levels.}
\label{Table:neural_results}
\end{table*}

\paragraph{Multi-Level Attention for BERT.}

As noted by \citet{ethayarajh2019contextual} the deeper BERT goes, the more ``contextualized'' its representation becomes. However, interpreting semantics of entities requires contextual knowledge in different degrees and always taking the last layer's output may not be the best solution. In order to address this issue, we propose a Multi-Level Attention~(denoted as BERT\nobreakdash-term$_\text{MLA}$) module on top of BERT to further enhance the representation extracted from BERT by learning how much to attend to each layer for producing an entity representation.
%
%
The attention weights of the $i$-th layer is computed as $a_i = [\ \mathbf{B}_i \cdot \mathbf{A} \ ]_+$, 
where $[\cdot]_+ = \max(0, \cdot)$, and $\mathbf{B}_i\in\mathbb{R}^{d}$ denotes the representation from the $i$-th level of BERT, $d$ denotes the dimensionality (i.e., here $d=768$), and $\mathbf{A}\in\mathbb{R}^d$ denotes a trainable attention memory vector.
We further normalise $a_i$ using a softmax layer, $w_i\stackrel{\text{def}}{=}\texttt{softmax}(a_i)$.
Finally, a weighted sum over all layers produces the attention-fused representation, i.e. $\text{BERT-term}_{\text{MLA}}=\sum_i^{L} w_i \mathbf{B}_i$.

\paragraph{Concept Embeddings.} We experimented by embedding SNOMED concepts with two modalities: (i) their labels, to exploit textual information, and (ii) their corresponding nodes in the KG, to incorporate the graph structure. Label embeddings were produced by running FastText (denoted as FT-label) and BERT (denoted as BERT-label) on the label, both trained as described above; for concepts with multiple labels (e.g., 
SCTID: 61685007 - \emph{Lower extremity}, \emph{Lower limb}, \emph{Leg}),
the mean of the label representations is used. For node embeddings, we based our choice of model on the findings reported in \citet{agarwal2019snomed2vec} and opted for their best reported model for SNOMED, i.e. node2vec~\cite{grover2016node2vec} with the suggested parameters and vector size 300.\footnote{We also compared node2vec with more sophisticated model of \citet{kartsaklis2018mapping} but we observed worse performance. We speculate this is due to the reliance of their model on the presence of textual definitions in SNOMED labels, which is only available in $<4\%$ of SNOMED nodes.} 

\paragraph{Ensemble Embeddings.} 
We also considered several embeddings that integrate multiple views of the data via (i) concatenation (denoted as $\oplus$) of the entity embeddings (e.g, FT-term $\oplus$ $\text{BERT-term}_{\text{MLA}}$), and (ii) concatenation of label and node2vec embeddings for concepts (e.g.,  FT-label $\oplus$  BERT-label $\oplus$ node2vec).

\paragraph{Alignment Model.} We adopt a linear transformation followed by ReLU \citep{nair2010rectified} 
for aligning entity and concept embeddings, and
we train the model with a max-margin triplet loss:
\begin{equation}
\mathcal{L} = \sum_{\textbf{p}\in \mathcal{P}} \max_{\bar{\textbf{t}}\in \mathcal{T} \backslash \{\textbf{t}\}} [\ \alpha -s(\textbf{p},\textbf{t}) + s(\textbf{p},\bar{\textbf{t}})\ ]_+ 
\label{eq:loss}
\end{equation}
\noindent where $\alpha$~($=0.2$) is a pre-set margin, $s(\cdot,\cdot)$ is the cosine similarity, $\mathcal{P}$ and $\mathcal{T}$ are the sets of all predictions and target embeddings in a mini-batch, and given a prediction $\textbf{p}$ and its corresponding ground truth $\textbf{t}$, $\bar{\textbf{t}}$ denotes a negative target embedding.
 
\paragraph{Results.} The results of the neural baselines are presented in Table~\ref{Table:neural_results}. All individual baselines (n.1 to n.4) fall behind the string-matching methods on Acc@1. This can be due the fact that on average for each entity-concept pair there are less than 4 examples
even in the stratified training set, making it difficult for the trained model to generalise well. This issue is 
more evident in the zero-shot setting.

The ensemble neural baselines
compensate for the lack of training signal by leveraging multiple views of the data. As expected, combining both surface and node embeddings of the concepts (n.5) offers a slight improvement, but still fails to match the string-matching baselines. Finally, concatenation of the entity embeddings with our proposed BERT-term$_\text{MLA}$ representation, and of the label embeddings with BERT-label (n.6) outperforms all previous baselines on the stratified split, but still falls behind the string-based baselines on zero-shot. 

Compared to Acc@1, while the overall ranking of models remains the same, MRR and Acc@10 are more forgiving.
The significant gap between Acc@1 and Acc@10 suggests that a re-ranking step~\cite{10.1561/1500000016} applied to top-10 candidates could further boost the performance. We leave further exploration of this idea to our future work.

 
\subsection{Back-off Baselines} 
\label{sec:back-off}
To obtain the best possible performance, we experimented with a deterministic back-off procedure (denoted as $+$) that applies the Dictionary and backs-off to a String-Matching model (\cref{sec:string-dictionary}) and finally to the best ensemble model (\cref{sec:neural-baselines}; model n.6 in Table~\ref{Table:neural_results}) for handling the missed cases. 

\paragraph{Results.}
Table~\ref{Table:backoff_results}
reports the Back-off baseline results. The immediate gain on performance compared to each individual counterpart indicates that each model is equipped to tackle only a subset of the underlying challenges in the data. The back-off model combining dictionary, Stoilos distance, and the ensemble neural approach achieves our best performance across both splits (model b.8 in Table~\ref{Table:backoff_results}). As expected, the neural baselines contribute much less in the Zero-Shot split with a meagre 4\%(3\%) improvement, compared to the 8\%(7\%) increase on the Stratified split. 
\begin{table}
\small
\setlength{\tabcolsep}{5.8pt}
\centering
\begin{tabular}{lccc}
\toprule
&&\multicolumn{2}{c}{{Acc@1}}\\
\cmidrule[1.5pt]{3-4}
\# & Method &Stratified Split & Zero-Shot Split\\
\midrule
b.1 & s.1 $+$ s.2 & .66 (.59) & .37 (.35) \\
b.2 & s.1 $+$ s.3 & .70 (.64) & .52 (.49) \\
b.3 & s.1 $+$ s.4 & .71 (.65) & .53 (.51) \\
\midrule
b.4&s.1 $+$ n.6&.77 (.70)& .36 (.33)\\
b.5&s.2 $+$ n.6&.71 (.67)&.53 (.49)\\
b.6&s.1 $+$ s.2 $+$ n.6&\textbf{.79 (.73)}&.53 (.49)\\
b.7&s.1 $+$ s.3 $+$ n.6&\textbf{.79} (.72)&.56 (.53)\\
b.8&s.1 $+$ s.4 $+$ n.6&\textbf{.79} (.72)&\textbf{.57 (.54)}\\
\bottomrule
\end{tabular}
\caption{Back-off baselines on stratified and zero-shot splits for general and (specific) levels.}
\label{Table:backoff_results}
\end{table}
Even if their overall contribution is limited, we were able to verify that our neural baselines are actually able to exploit the context as expected. For example w.r.t. the issues typical of the UGC domain we identified in Section~\ref{sec:intro}, we found neural methods helpful in resolving 
acronyms (
``UTIs'' to SCTID: 68566005 -- \emph{Urinary Tract Infection}), 
colloquial synonyms (
``bloodwork'' to SCTID: 396550006 -- \emph{Blood Test}), 
compositionality (
``drenched in sweat'' to SCTID: 415690000 -- \emph{Sweating}), complex inference 
(e.g., ``Oral Cancer" to SCTID: 363505006 -- \emph{Malignant tumour of oral cavity}), or even spelling errors combined with alternative product names (``Remicaid'' to SCTID: 386891004 -- \emph{Infliximab}, i.e. the active principle of \emph{Remicade}). This last example is specifically interesting, since the label \emph{Remicade} is not present in SNOMED but the pre-training of embeddings on medical texts (\cref{sec:neural-baselines}) allowed the neural baselines to pick up the correct node.

\section{Discussion}
\label{sec:discussion}
The COMETA corpus introduces a challenging scenario for entity linking systems from both ML and NLP perspectives. In this section we summarise these challenges, our findings, and shed light on aspects that demand future attention:

\paragraph{Domain-Specific Language.} EL systems similar to our baselines are not uncommon in the biomedical domain: \citet{furrer-etal-2019-uzh} used a similar dictionary-BERT ensemble model to achieve the best performance in the 2019 CRAFT Shared Task~\cite{baumgartner-etal-2019-craft} on biomedical literature. However, in their case, the neural component offered a much higher contribution highlighting the underlying challenges in medical layman's language. Additionally, probing our proposed Multi-Level Attention for BERT, we observed that a more flexible utilisation of context is effective in understanding the diverse contextual cues. 

\paragraph{Low-Resource Regime and Learning.}
Compared to similar corpora, COMETA has the largest scale. However, from a learning perspective the lack of sufficient regularity in the data could still leave its toll at test phase. This is a natural consequence of high productivity of layman's language in social media, while emerging and unforeseen topics such as pandemics (i.e., COVID19) could also contribute to the problem. In fact, we observed the daunting task that systems face in the zero-shot setting, where in the absence of sufficient training signal, string-based methods offer a strong baseline which is hard to beat for neural counterparts. While we artificially control this in the stratified split we still believe the zero-shot setting draws a more detailed picture of challenges an EL system needs to tackle in a real-world scenario. Further exploration of solutions such as transfer learning across domains (i.e., from medical literature to layman's domain) is beyond the focus of this work, nonetheless COMETA provides the framework for designing and testing such solutions.

\paragraph{Cross-Modality Alignment.} While \citet{agarwal2019snomed2vec} report superior performance of node2vec  embeddings on several graph-based tasks on SNOMED, this success does not translate into EL as it relies on mapping across modalities (i.e., text-to-graph). Alternatively, when we replaced the node2vec with concept-label embeddings (produced by FT/BERT) the performance was significantly improved. This suggests that aligning different modalities may require a more complex alignment model or stronger training signals. We leave further exploration of this to future work.

\section{Conclusion}
\label{sec:conclusion}
We presented COMETA, a unique corpus for its scale and coverage which is curated to maintain high quality annotations of medical terms in layman's language on Reddit with concepts from SNOMED knowledge graph. Different evaluation scenarios were designed to compare the performance of conventional dictionary/string-matching techniques against the mainstream neural counterparts and revealed that these models complement each other very well and the best performance is achieved by combining these paradigms. Nonetheless, the missing performance of 28-46\% (depending on the evaluation scenario) encourages future research on this area to take this corpus as a challenging yet reliable evaluation benchmark for further development of models specific to this domain.



COMETA is available by contacting the last author via e-mail or following the instructions on \url{https://www.siphs.org/}.
We release the pre-trained 
embeddings and the code to replicate our baselines online at \url{https://github.com/cambridgeltl/cometa}.

\section{Acknowledgments}

Funding: This work was supported by the UK EPSRC (EP/M005089/1). 
We kindly acknowledge Molecular Connections Pvt. Ltd\footnote{\url{http://www.molecularconnections.com/}} for their work on annotating our data.

\bibliography{bibliography}
\bibliographystyle{acl_natbib}

\clearpage

\appendix
\label{sec:appendix}
\section{Appendices}

\subsection{Full List of Subreddits}\label{fullistreddits}
Table~\ref{Table:68} reports the list of 68 subreddits crawled for COMETA.
\begin{table*}
\centering
\begin{tabular}{p{3cm}p{3cm}p{3cm}p{3cm}}
\toprule
healthIT          & hepc          & Cirrhosis          & breastcancer       \\
AskDocs           & T1D           & scoliosis          & Colic              \\
DiagnoseMe        & diabetes      & health             & PsoriaticArthritis \\
cancer            & Constipated   & cfs                & Thritis            \\
ChronicPain       & Constipation  & DuaneSyndrome      & fibro              \\
dementia          & migraine      & atrialfibrillation & HiatalHernia       \\
flu               & panicdisorder & insomnia           & PCOS               \\
mentalhealth      & benzorecovery & DSPD               & Urology            \\
MultipleSclerosis & Psoriasis     & braincancer        & multiplemyeloma    \\
STD               & ClotSurvivors & Hypermobility      & leukemia           \\
transplant        & rheumatoid    & GERD               & lymphoma           \\
birthcontrol      & Sciatica      & seizures           & AskaPharmacist     \\
menstruation      & urticaria     & dialysis           & mastcelldisease    \\
antidepressants   & crazyitch     & ChronicIllness     & obgyn              \\
Allergies         & pancreatitis  & askdentists        & askadentist        \\
FoodAllergies     & CrohnsDisease & Dentistry          & HealthInsurance    \\
Allergy           & Ovariancancer & Antibiotics        & hearing            \\ \bottomrule
\end{tabular}
\caption{The list of the 68 subreddits used as a source for the corpus.}
\label{Table:68}
\end{table*}

\subsection{Example from COMETA}\label{examplefromcometa}
Table~\ref{table:example-dataset} provides examples from COMETA and illustrates the structure of each line in the corpus.
\begin{table*}
\setlength{\tabcolsep}{3.5pt}
\begin{tabular}{ p{1cm}p{1cm}p{2.8cm}p{2.8cm}p{4cm}p{2.8cm}  }
 \toprule
 \textbf{ID}&\textbf{Term}&\textbf{General SCTID}&\textbf{Specific SCTID}& \textbf{Example}&\textbf{Subreddit} \\
 \texttt{int} &\texttt{str} &\texttt{int} &\texttt{int} &\texttt{str} &\texttt{str} \\
 \midrule
\dots &\dots &\dots &\dots &\dots &\dots \\
$i$   & acid & 34957004                & 34957004              & I burned myself with acid  & AskDocs \\
$i+1$ & acid & 34957004                & 698065002              & acid in my throat & cancer \\
\dots &\dots &\dots &\dots &\dots &\dots \\
 \bottomrule
\end{tabular}
\caption{The structure of the dataset; column names are denoted by \textbf{bold} text, and column types are denoted by \texttt{monospaced} text. The released dataset contains two additional columns, marking the label for the corresponding General and Specific SCTID respectively. However, since a label may appear in multiple nodes, we recommend to \emph{always} use SCTIDs to retrieve the target nodes.   
\\Please note that the data in this table is used for illustration purposes only and it might not be contained in the released corpus.}
\label{table:example-dataset}
\end{table*}

\subsection{Example from Assessor Guidelines}\label{examplefromassessorguideline}
Table~\ref{table:assessor} provides an example from the guideline sent to assessors.
\begin{table*}
\setlength{\tabcolsep}{3.5pt}
\begin{tabular}{{
p{1.8cm} 
p{4cm } 
>{\raggedright\arraybackslash}p{2.2cm} 
>{\raggedright\arraybackslash}p{3cm } 
p{3.5cm}
}}
\toprule
{\bf Quality}&{\bf Evaluation}&{\bf Term}&{\bf Proposed Node}&{\bf Explanation}\\
\midrule
5:Excellent&The SNOMED node \textbf{matches exactly} the term or is a \textbf{synonym} of the term.&Chronic back pain&\href{https://browser.ihtsdotools.org/?perspective=full&conceptId1=134407002&edition=MAIN/2020-03-09&release=&languages=en}{Chronic  back pain, 134407002}&	Exact match.\\
\midrule
4:Good&The SNOMED node is \textbf{conceptually similar and taxonomically close} (1-2 edges) to the target term, e.g. is a close ancestor/descendant or a sibling.&	Chronic back pain&\href{https://browser.ihtsdotools.org/?perspective=full&conceptId1=161891005&edition=MAIN/2020-03-09&release=&languages=en}{Back pain, 161891005}&`Back pain' is the direct ancestor of `Chronic back pain'.\\
\midrule
3:Fair&The SNOMED node is \textbf{conceptually related} and reasonably close (1 to 3 edges) to the target term, both taxonomically or via attributes (finding site, etc.)&Chronic back pain&\href{https://browser.ihtsdotools.org/?perspective=full&conceptId1=77568009&edition=MAIN/2020-03-09&release=&languages=en}{Back, 77568009}&`back' is the `finding site' of `Chronic back pain'.\\
\midrule
2:Poor&The SNOMED node is \textbf{conceptually distant} from the term, and there is a reasonably long (3-4 edges) path from it to the correct node&Chronic back pain&\href{https://browser.ihtsdotools.org/?perspective=full&conceptId1=22943007&edition=MAIN/2020-03-09&release=&languages=en}{Torso, 22943007}&	 `Chronic Back Pain' is located in the `Torso', so they are somewhat related, and the two nodes are not far (distance 3)\\
\midrule
1:Very Poor&The SNOMED node is \textbf{completely unrelated} with the term, and the path between the correct node and the target one is very long ($>5$).&Chronic back pain&\href{https://browser.ihtsdotools.org/?perspective=full&conceptId1=61968008&edition=MAIN/2020-03-09&release=&languages=en}{Syringe, 61968008}& `Chronic Back Pain' and `Syringe' have high distance (5), \textbf{and} the concepts are completely unrelated. \\
\bottomrule
\end{tabular}
\caption{An example from assessor guidelines.}
\label{table:assessor}
\end{table*}

\subsection{Distribution of Concepts in Stratified and Zero-Shot Splits}\label{distributionofconceptsinstratifiedandzeroshot}
\begin{figure*}[b]
  \centering
    \includegraphics[width=0.66\textwidth]{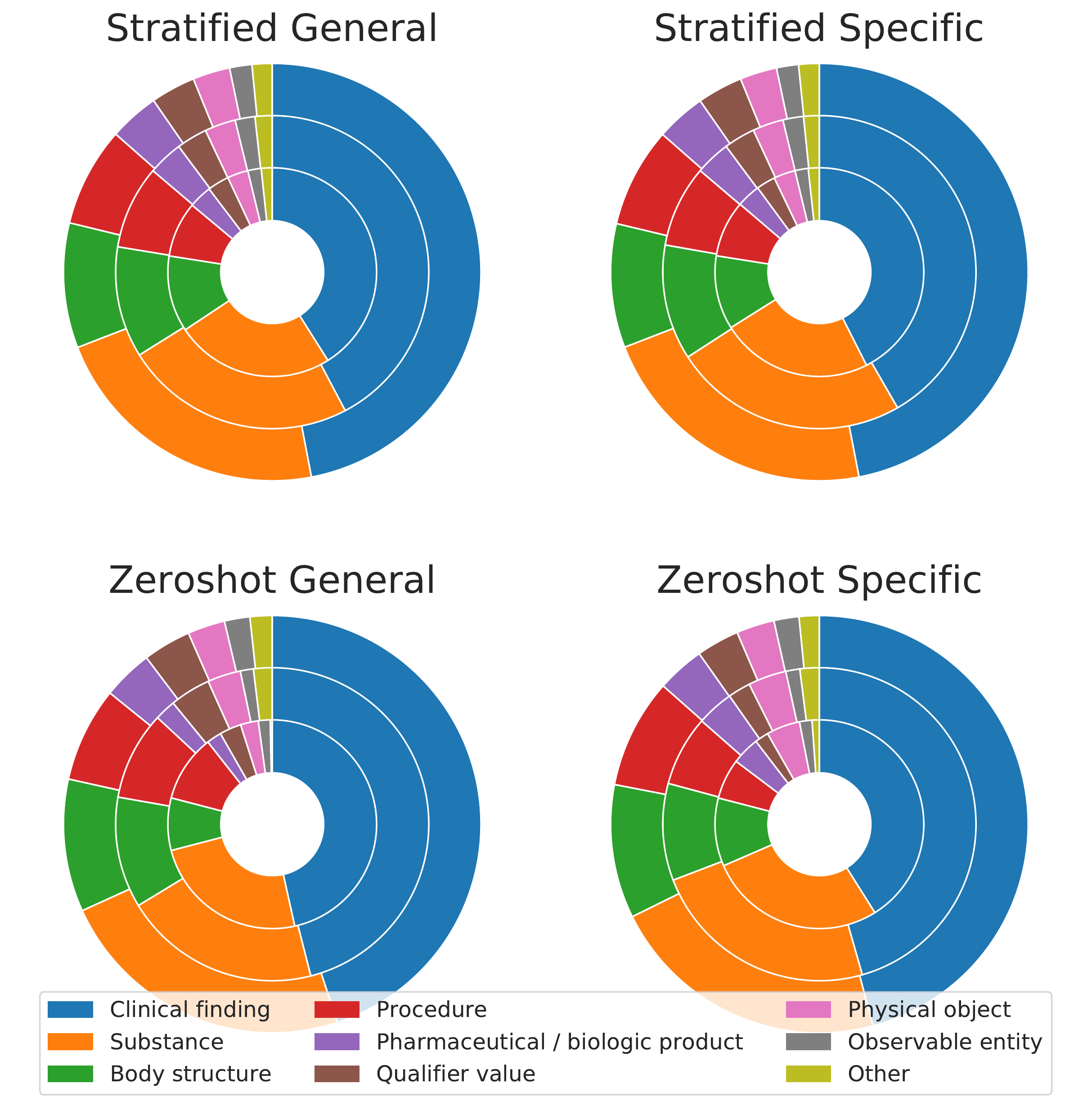}
    \caption{The categories in the dataset by split. The outer pie is the training set, the middle pie is the test set, the inner pie is the development set.}
    \label{fig:detailed_distribution}
\end{figure*}

Figure~\ref{fig:detailed_distribution} provides the detailed distribution of SNOMED Concepts in Stratified and Zero-Shot splits.

\subsection{Reproducibility}\label{reproducibilitychecklist}
\Cref{Table:hardware} and \Cref{Table:search_space} describe the hardware and hyperparameters used for the experiments we describe.

\subsection{Stoilos Distance}\label{stoilosdistancedefinition}
The commonality function $\texttt{comm}(x,y)$,  is defined as 
\begin{equation*}
    \texttt{comm}(x,y) = \frac{ 2\cdot \sum_{i}|\text{max\_common\_substring}|}{(|x|+|y|)/2}
\end{equation*}
Where the max\_common\_substring between $x,y$ is computed in an iterative manner: first, that of the original $x,y$ are computed; then the common sub-string is removed and search is done again for the next max\_common\_substring until a threshold of length $3$ is met (common sub-strings with $<3$ length are not considered).

The difference function, $\texttt{diff}(x,y)$, is based on the unmatched part of $x,y$ from the last step. We denote them as $u_x,u_y$. And the length of them are normalised using a Hamacher product \citep{hamacher1978sensitivity} (a parametric triangular norm):
\begin{equation*}
{\displaystyle 
    \texttt{diff}(x,y) = \frac{|u_x|\!\cdot\!|u_y|}{p\! +\! (1\!-\!p)\!\ \! (|u_x|\!+\!|u_y|\! -\! |u_x|\!\cdot\!|u_y|)}
}
\end{equation*}
We choose $p=0.6$.

\begin{table*}[t!] 
\small
\setlength{\tabcolsep}{1pt}
\centering
\begin{tabular}{lr}
\toprule
hardware & specification \\
\midrule
 RAM & 64 GB \\
 CPU & AMD\textsuperscript{\textregistered} Ryzen 9 3900x 12-core 24-thread \\
 GPU & NVIDIA\textsuperscript{\textregistered} GeForce RTX 2080 Ti (11 GB) $\times$ 2\\
\bottomrule
\end{tabular}
\caption{Hardware specifications of the machine used to run our experiments.}
\label{Table:hardware}
\end{table*}

\begin{table*}[t!] 
\small
\setlength{\tabcolsep}{1pt}
\centering
\begin{tabular}{lr}
\toprule
hyper-parameters & search space \\
\midrule
optimiser & \{AdamW$^\ast$, Adam\} \\
learning rate & \{\texttt{1e-4}$^\ast$, \texttt{5e-4}, \texttt{1e-5}$^\dagger$\} \\
batch size & \{64$^\ast$, 128, 256\}\\
training epochs & \{30, 50$^\ast$, 100\} \\
$\alpha$ in Eq. (1) & \{0.05, 0.1, 0.2$^\ast$\}\\
threshold for Levenshtein (b.7) & [0.10, 0.20] \\
threshold for Stoilos (b.8) & [0.05, 0.10] \\
BERT pre-training global step & \{10k, 100k$^\ast$\}\\
BERT pre-training max\_seq\_length & \{64$^\ast$, 128\}\\
\bottomrule
\end{tabular}
\caption{This table lists the search space for hyper-parameters; $^\ast$ denotes the ones used to obtain the performance described in this publication if not specified otherwise. $^\dagger$ identifies parameters used only for models n.5 and n.6. More details can be found in the source code available online at \texttt{redacted}. Details of the two optimisers are specified in \citet{loshchilov2018decoupled} 
and \citet{kingma2015adam}.
}
\label{Table:search_space}
\end{table*}

\end{document}